# Semantic segmentation of forest stands using deep learning

Håkon Næss Sandum[a*], Hans Ole Ørka[a], Oliver Tomic[b], Erik Næsset[a] and Terje Gobakken[a]

[a] Faculty of Environmental Sciences and Natural Resource Management, Norwegian University of Life Sciences, NMBU, P.O. Box 5003, NO-1432 Ås, Norway.

[b] Faculty of Science and Technology, Norwegian University of Life Sciences, NMBU, P.O. Box 5003, NO-1432 Ås, Norway.

* hakon.nass.sandum@nmbu.no

Abstract

Forest stands are the fundamental units in forest management inventories, silviculture, and financial analysis within operational forestry. Over the past two decades, a common method for mapping stand borders has involved delineation through manual interpretation of stereographic aerial images. This is a time-consuming and subjective process, limiting operational efficiency and introducing inconsistencies. Substantial effort has been devoted to automating the process, using various algorithms together with aerial images and canopy height models constructed from airborne laser scanning (ALS) data, but manual interpretation remains the preferred method. Deep learning (DL) methods have demonstrated great potential in computer vision, yet their application to forest stand delineation remains unexplored in published research. This study presents a novel approach, framing stand delineation as a multiclass segmentation problem and applying a U-Net based DL framework. The model was trained and evaluated using multispectral images, ALS data, and an existing stand map created by an expert interpreter. Performance was assessed on independent data using overall accuracy, a standard metric for classification tasks that measures the proportions of correctly classified pixels. The model achieved an overall accuracy of 0.73. These results demonstrate strong potential for DL in automated stand delineation. However, a few key challenges were noted, especially for complex forest environments.

# 1. Introduction

A forest stand is a cohesive community of trees with sufficient consistency in attributes to distinguish it from adjacent communities. These attributes include species composition, structure, age, size class distribution, stocking level, spatial arrangement reflecting historical and local silvicultural practices, and site characteristics such as topography and site index (Baker, 1950; Husch et al., 1993; Smith, 1986). According to common practice in Norway, a stand typically covers a minimum area of 0.2 hectares, forming a relatively homogenous area suited for a specific management regime, and serving as the fundamental unit in inventory, for forest management, and financial analysis. However, stand boundaries are not always uniform or easily discernible (De Groeve & Lowell, 2001). Some stand boundaries are well-defined and easy to distinguish, such as crisp boundaries between mature forests and adjacent clear-cuts. In contrast, other boundaries are more ambiguous, especially when adjacent stands have only slight differences in characteristics. Additionally, some boundaries do not exist as sharp divisions in the landscape but as transition zones where forest characteristics gradually change.

Stand-level information aids forest managers in making decisions regarding silvicultural treatment. Certain important characteristics, such as timber volume, site index, and stem number, are calculated for each stand. These characteristics are often calculated on a per-area basis, making them sensitive to the area estimate of the individual stands. Errors in areal estimates of forest stands arise from incorrect boundary placement, where the manually interpreted stand boundaries differ from the real borders (Næsset, 1999a). Positional errors may have a considerable effect on estimates of properties when adjacent stands have substantial differences, such as the boundaries between mature forests and clear-cuts.

The stand delineation procedure is typically a somewhat subjective manual procedure based on spectral and structural information from aerial images, and has a long-standing history (Andrews, 1934). Photogrammetry, a technique that uses overlapping aerial images to obtain three-dimensional (3D) information, has been employed to assist in the stand delineation process, by improving depth perception and adding additional detail. For boreal forests, for example, the texture of the canopy and the spectral information provide insight into the dominant tree species and growing conditions (Axelson & Nilsson, 1993). However, the delineation process is not always straightforward. Shadows and tree branches can obscure the ground view in aerial images, making it difficult to accurately place stand boundaries leading to inaccuracies in border placement (Næsset, 1998). Studies applying simulations to the borders of single stands neighboring clearcuts found that shadows led to a 1.3% to 9.6% underestimation of the mature forest stand area (Næsset, 1999a), and a 2% underestimation of the area of thinning phase forest when bordering clear-cuts (Næsset, 1999b). Additionally, boundary placement can vary significantly between different interpreters analyzing the same area (Næsset, 1998), and even the same interpreter may produce different results when analyzing the same area on two different occasions (Nantel, 1993). This combination of inaccuracies and subjectivity, combined with the manual process being time-consuming, makes the problem lend itself as a good candidate for automation, and has resulted in several studies combining arial images and airborne laser scanning (ALS) data with various algorithms.

Multiple studies have applied aerial images for automated stand delineation, demonstrating their effectiveness as input data for automation (e.g. Leckie et al., 2003). An aerial image contains spectral and structural information valuable for stand delineation; however, lack perception of depth and do not adequately capture the forest's vertical structure. An alternative to aerial images is to construct a canopy height model (CHM) from ALS data. A CHM provides a comprehensive representation of forest structure, capturing size, shape, distribution, and height of tree crowns, which facilitates more accurate delineations (Mustonen et al., 2008). CHMs for automated stand segmentation have formed the basis for several publications (Eysn et al., 2012; Koch et al., 2009). While CHMs provide a good representation of forest structure, some areas lack distinguishing structural features, such as marshes and clear-cuts. These areas are more easily distinguished in spectral data. The lack of vertical information in aerial images and spectral information in CHM can be mitigated by integrating both data sources, as suggested by Diedershagen et al. (2004) and Hernando et al. (2012). Dechesne et al. (2017) obtained better results when combing the two data sources as opposed to applying them separately.

The most prominent approach in the published literature is the combination of remotely sensed data with the eCognition software (Trimble Germany GmbH, 2021) and it has shown promising results (e.g. Hernando et al., 2012). Applying the region-based algorithm by Baatz and Schäpe (2000), polygons are created by growing regions from initial seeds based on user defined initial parameters. However, the algorithm is sensitive to the settings of these parameters (Mustonen et al., 2008). To overcome this limitation, a combination of trial and error and knowledge of local forest characteristics may be required for parameter tuning and does not represent a fully automated procedure.

Variability in background, lighting, object orientation and location, and image quality introduces inconsistencies that can hinder the reliability of automated methods, making it difficult for models to consistently and accurately delineate forest stands. The introduction of deep learning (DL) models in the field of computer vision has shown promise in overcoming such challenges. However, the application of DL for stand delineation remains unexplored in scientific literature. This study aims to address this gap by applying a DL model to the stand delineation process.

Among the most successful DL approaches in image processing are convolutional neural networks (CNNs), which extract hierarchical features through convolution and pooling layers. While CNNs are highly effective for image classification tasks, they cannot precisely locate the extracted features, making the insufficient for segmentation purposes, which require precise location.

To address this limitation of spatial information loss, Ronneberger et al. (2015) proposed the U-Net architecture, a specialized segmentation model. U-Net builds on the CNN framework but applies a symmetric encoder-decoder structure. The encoder, also known as the contracting path of the model, operates similarly to traditional CNNs, reducing the image dimensions while capturing relevant features. The decoder, also known as the expansive path, restores spatial information by using transposed convolutions and skip connections. The encoder-decoder structure ensures that the final output retains both the semantic understanding of the image and the precise spatial location of the objects, and allows the model to produce high-quality, pixel-wise segmentation results.

Several recent studies have already applied U-Net-like models to segment remotely sensed data for natural resource mapping and forestry applications. Promising results were found for different use cases, such as mapping of raised bog communities (Bhatnagar et al., 2020), segmentation of tree species (Kentsch et al., 2020; Schiefer et al., 2020), and mapping of wheel ruts after harvest (Bhatnagar et al., 2022). The model's relative simplicity should also make it a good starting point for the introducing DL to stand delineation.

The objective of this study was to develop an automated forest stand delineation procedure that can substantially reduce the time and labor requirements, while also minimizing subjectivity. To achieve this, we propose a novel approach to the delineation problem, framing it as a multiclass segmentation problem that integrates multispectral aerial images, a CHM, and U-Net, a DL model known for its ability to deliver strong segmentation results even with limited data. By employing this supervised DL framework, the model is designed to replicate human interpretations and capture nuanced patterns informed by experience and cultural context. An independent dataset was used to evaluate the model's performance, with assessment based on metrics describing the level of agreement between the model predictions and the reference data. In addition, visual inspection was conducted to account for the fact that multiple valid realizations of stand boundaries are possible to achieve for a given area.

## 2. Materials & Methods

### 2.1 Study area

The study was conducted using data from a large private forest property spanning 358 km$^2$ in Akershus county, approximately 50 km north of Oslo (Figure 1). The property, owned by Mathiesen Eidsvold Værk ANS (MEV), is actively managed and certified under both the Forest Stewardship Council (FSC) and the Program for the Endorsement of Forest Certification (PEFC) standards. In accordance with the FSC guidelines, 5% of the property is designated as protected, with another 5% being subject to specific harvesting constraints (Løvli, 2022). The forest is dominated by Norway spruce (*Picea abies* (L.) Karst.), with smaller areas of Scots pine *(Pinus sylvestris* L.*)* and broadleaved species. The property's elevation ranges from 176 to 812 m above sea level. The most recent forest inventory for forest management planning was conducted in 2021.

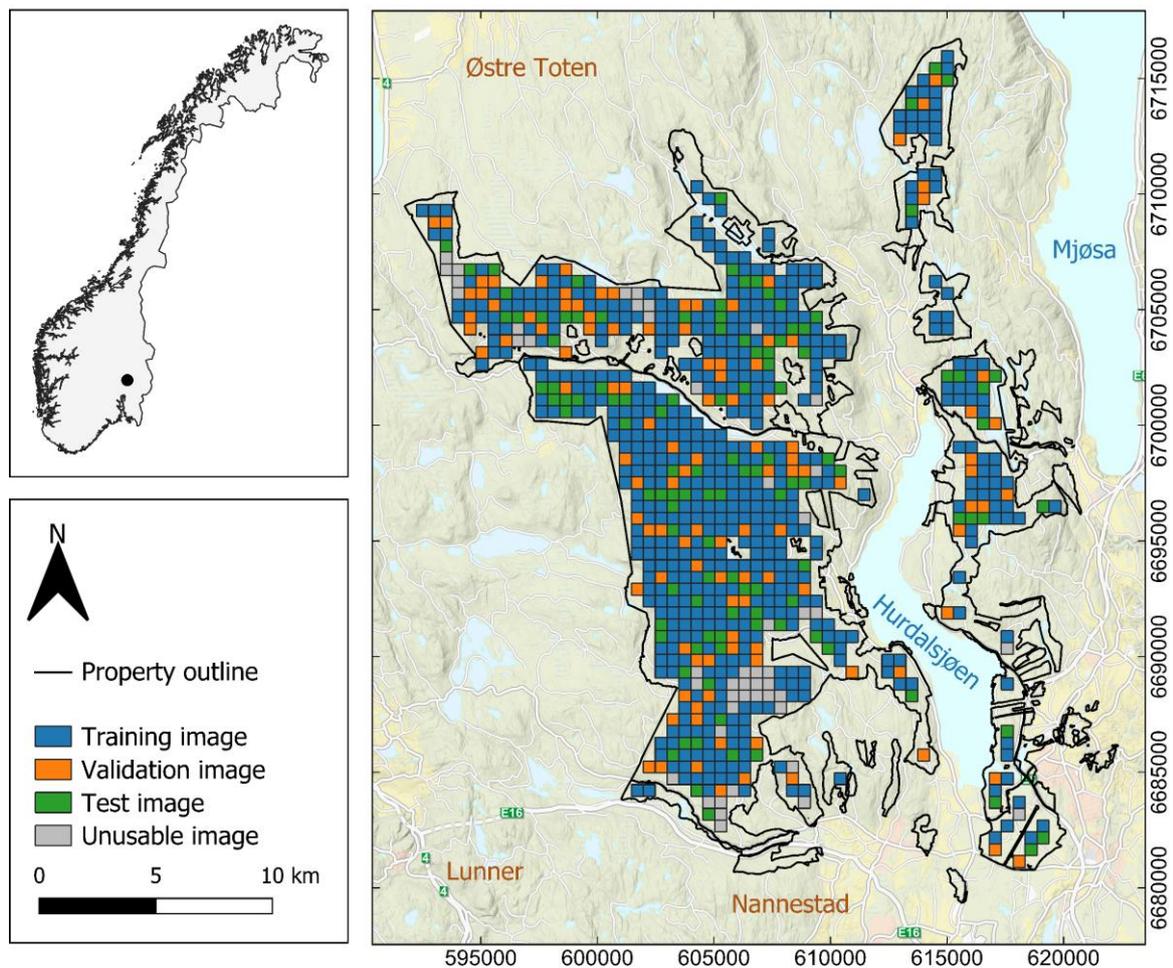

*Figure 1. Map of the forests of Mathiesen Eidsvold Værk ANS located in Akershus County, Norway. The area was divided into images tiles of 512 m × 512 m.*

## 2.2 Datasets

In this study, three primary datasets were utilized: multispectral aerial images, ALS data, and a stand map created by an expert interpreter with 30 years of experience. These datasets form the foundation of the U-Net modelling process for delineating forest stands. The aerial images provide spectral information, the ALS data supports the construction of a CHM, and the reference data serves as the basis for creating masks for training and validating the model. Each dataset is described in detail below.

### 2.2.1 Aerial Images

Aerial images were obtained from Norway's national aerial image database (https://www.norgeibilder.no/). Under Norway's national aerial image acquisition program, images are captured every 5-10 years, depending on the type of area (Norwegian mapping authority, 2024), and are stored in the database as orthophotos. The images used in the current study were acquired on three different dates, August 13th, August 14th, and September 1st, 2022, from an altitude of 6300 m above ground level. The images contain four color channels – red, green, blue, and near-infrared – with a spatial resolution of 0.25 m, and an 8-bit radiometric resolution.

### 2.2.2 ALS data

ALS data was collected on July 17th, 2021, as part of the forest management planning inventory. A dual channel Riegl VQ 1560II was flown at 3500 m above ground level, with a pulse repetition rate of 250 kHz per channel, a scan rate of 56 lines per second, and a 20% lateral overlap between lines. The resulting pulse density was 1.4 pulses $m^{-2}$.

### 2.2.3 Reference data

In DL, the term "ground truth" typically refers to the data used for model training and evaluation. However, this poorly reflects the highly subjective nature of stand delineation and the interpretive process of creating maps (Magnusson et al., 2007). Therefore, the manually delineated stands are referred to as "reference data".

During the 2021 forest management planning inventory, 13,000 stands were manually delineated using photogrammetry with a CHM constructed from the ALS data as support (Table 1). Each stand was classified as either forest or non-forest. The forest stands were further attributed with additional information, such as species composition, age, site index, and stand development stage. Development stage was determined based on age, species, and site index. Stage I includes bare forest land, while Stage II represents recently regenerated forest with trees up to 8-9 m in height. Stage III is designated for young production forests, Stage IV for old production forests, and Stage V for mature forests (Anon, 1987). This division aids forest managers in making informed decisions regarding silvicultural treatments. Stage I represents forests in need of regeneration, while Stage II indicates areas where pre-commercial thinning should be considered. Stages III and IV are young and old thinning-phase forests, and Stage V consists of mature forests ready for harvest. These development stages served as the basis for creating reference masks used in model training and evaluation.

Table 1. Forest stand characteristics from the stand database created during the 2021 forest management planning inventory.

| Class[1] | Stands (n) | Area (ha) | Mean age | Volume (%) | | | Lorey's mean height (m) | | |
|---|---|---|---|---|---|---|---|---|---|
| | | | | Spruce | Pine | Broadleaves | Min | Max | Mean |
| NF | 3079 | 5159 | - | - | - | - | - | - | - |
| I | 260 | 538 | 0 | - | - | - | 0.0 | 22.0 | 0.4 |
| II | 1966 | 4799 | 15 | - | - | - | 0.1 | 20.0 | 2.3 |
| III | 2828 | 10803 | 44 | 90 | 2 | 8 | 5.1 | 23.3 | 13.4 |
| IV | 2302 | 6563 | 68 | 89 | 2 | 9 | 6.5 | 25.4 | 16.2 |
| V | 2889 | 7938 | 101 | 91 | 3 | 6 | 6.4 | 28.5 | 18.7 |

[1] NF – non-forest, I-V – stand development stages.

### 2.3 Pre-processing

Before segmentation can be performed, the data must undergo several preprocessing steps to ensure consistency and compatibility with the model. This section describes the preparation of the aerial images and ALS-data. Additional steps, including tiling, data cleaning, normalization, and the division of the data into training, validation and test set, are also outlined. Furthermore, the reference data used for modelling and evaluation, along with data augmentation techniques designed to enhance model robustness, are discussed. Together, these steps establish a dataset that is well-suited for segmentation.

### 2.3.1 Aerial images

When applying an image for modeling, a choice must be made regarding the spatial resolution of the image. An image with a finer spatial resolution provides the model with more detailed information but also requires longer processing time and larger memory usage. Additionally, with pixel-wise predictions, the spatial resolution of the image will affect the precision of the final delineation. A 1-m resolution was considered a suitable compromise between the level of information contained in the images, processing time, and the precision of stand boundaries produced by the model. Thus, the original 0.25-m resolution was downsampled to 1 m by averaging groups of four pixels.

### 2.3.2 ALS data

A CHM was constructed from the ALS data using the lidR v. 4.1.2 package (Roussel & Auty, 2024; Roussel et al., 2020) in R (R Core Team, 2023). It was created at a 1-m spatial resolution to match the aerial images and align with the previously mentioned considerations. The CHM was then incorporated as an additional channel to the aerial images, resulting in a five-layer raster composite.

### 2.3.3 Tiling, data cleaning, data split, and normalization

The raster composite was too large to fit into memory, and the model requires consistently sized image tiles. To address this, the raster composite was divided into tiles measuring 512 m × 512 m, producing individual 5-channel images of 512 × 512 pixels. Any images extending beyond the forest property boundaries were discarded to avoid issues with missing values. After this filtering, 809 images were retained. They were then visually inspected for discrepancies caused by the one-year time difference between the ALS and aerial image acquisitions. Twenty-eight images were excluded due to visible clear-cuts in the aerial images not reflected in the CHM, though minor discrepancies were permitted to avoid excluding and excessive number of images. Furthermore, images covering the protected areas with missing delineations were removed, leaving 760 images for analysis.

These images were randomly split into training (70 %, 532 images), validation (15 %, 114 images), and test (15 %, 114 images) sets. All images were normalized to a range of [0, 1] by dividing each pixel in the spectral layers by 255, and each pixel in the CHM with by 39, which corresponds to the largest observed value in the CHM.

### 2.3.4 Generating reference masks for segmentation

The U-Net model expects input in raster format, thus, to use the stand database as reference data for the model, the vector data was first rasterized based on the development stage of the forest. Due to the definition of the stand development stages I and II, these stages had to be combined. Typically, a stand is moved from development stage I to development stage II when the regeneration is deemed satisfactory, usually after planting. However, these plants are not visible in the spectral data and the CHM, necessitating the merging of development stage I and II. Ultimately, the raster layer representing the reference stands consisted of five classes: four based on development stages – Class I-II, Class III, Class IV, and Class V – and a single class for non-forested areas, Class NF. The raster was then one-hot encoded, a process that involves converting each class into a separate binary layer, with pixel values 1 indicating presence and 0 indicating absence of the corresponding class. This process created a five-layer raster, with one layer per class, referred to as masks. The distribution of the classes across the three datasets used for analysis is shown in Figure 2.

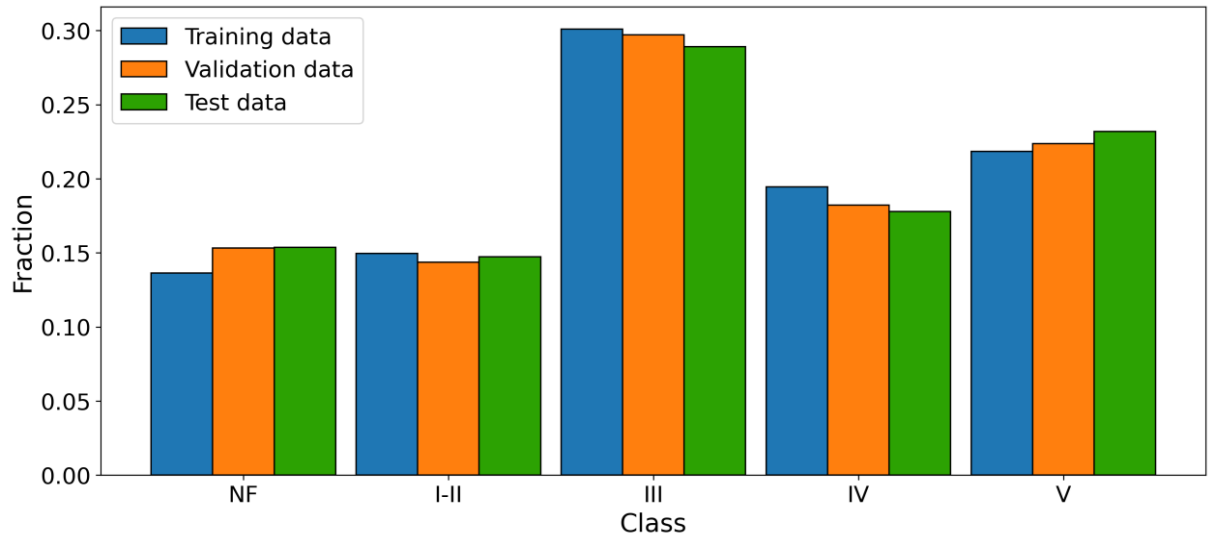

*Figure 2. Fraction of pixels across the five classes (NF – non-forest, I-II – V development stages) for each of the datasets used in model development and evaluation.*

### 2.3.5 Image augmentation

Training DL models requires large and diverse datasets to ensure that the model learns effectively and can generalize well to new, unseen data (Barbedo, 2018). Data augmentation, which involves modifying the appearance of images, is a technique used to artificially expand training datasets and improve performance in real-world applications (Goodfellow et al., 2016). In this study, common augmentation techniques, including horizontal flipping (e.g. Xu et al., 2023), brightness and contrast adjustments (e.g. Xiao et al., 2024) and Gaussian-noise addition (e.g. Moradi et al., 2020) were applied during model training, meaning the images were automatically altered between training epochs.

Shadows can affect the accuracy of the border placement between clear-cuts and mature forests (Næsset, 1998). The images used in this study were taken late in the season, with only three acquisition dates, and showed consistent shadow length and orientation. To account for real-world variability in shadow orientation caused by different acquisition times, horizontal flipping was used. This technique effectively reverses shadow orientation.

Brightness and contrast were adjusted with ±10% to simulate different lighting conditions. This value was chosen based on visual inspection and testing. This range ensured realistic representation without overly dark or bright outputs that could impair learning.

Gaussian noise was added to the images to reduce the model's reliance on fine detail, helping the model to learn more robust features and improve generalization (Sietsma & Dow, 1991). A zero-centered Gaussian distribution with a 0.1 standard deviation was used to balance challenge and realism, avoiding overly noisy images.

Initial tests helped determine the values of the augmentations discussed above. Smaller values tended to stabilize model convergence during later training stages, while larger values negatively impacted overall performance. Additionally, given the high-quality nature of the commercial aerial images from the national database, which are under strict quality control, there is minimal need to alter their appearance substantially, as the model is expected to encounter similarly high-quality images in real-world applications.

## 2.4 Model architecture

In this study, the general architecture of the U-Net (Ronneberger et al., 2015) was applied with slight adaptations. The model architecture is depicted in Figure 3, and the subsequent sections outline the key architectural components and modifications, including batch normalization, pooling operations, dropout regularization, weight initialization, activation functions, and the choice of loss function. These elements were carefully chosen to enhance the model's performance for stand delineation.

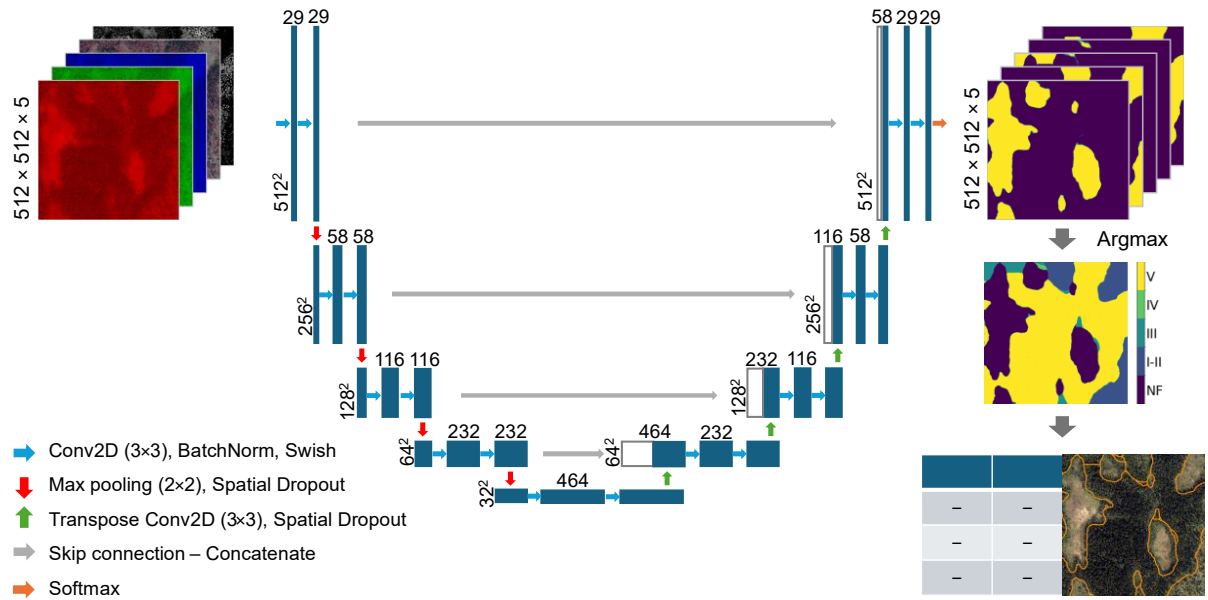

*Figure 3. Model architecture and flow of images through the model. Each blue box represents multi-channel feature maps produced by the model. The numbers along the vertical axis indicate the spatial dimensions of the images. The number of channels in the feature maps are denoted by the number on top of each box (After Ronneberger et al., 2015).*

### 2.4.1 Batch normalization

During training, the distribution of activations in DL models can shift as the parameters of previous layers update (Ioffe & Szegedy, 2015). Batch normalization mitigates this issue through normalization of layer inputs to have zero mean and unit variance. This leads to improved training stability and faster model convergence. Consequently, batch normalization was implemented in the model.

### 2.4.2 Pooling

Pooling operations were performed by using 2×2 max-pooling with a stride of 2. This procedure selects the strongest response within each 2×2 region of the feature maps produced by the convolutional layers. Selecting the strongest responses introduces robustness by retaining dominant features against noise and small variations (Matoba et al., 2022; Raschka & Mirjalili, 2019). Pooling also reduces the dimensionality of the feature maps, resulting in faster and more efficient calculations. Additionally, polling enables deeper layers to aggregate information over a larger spatial context, improving the model's ability to recognize patterns across varying positions (Nirthika et al., 2022).

### 2.4.3 2D dropout

In DL, a neuron refers to a computational unit within a neural network that processes input data and passes its output to subsequent layers. Dropping neurons in a procedure known as dropout is a common way to regularize neural networks as this reduces overfitting by preventing the model from relying too much on specific features. However, in convolutional layers, standard dropout is less effective because the surrounding pixels are spatially correlated, meaning that dropping individual neurons does not sufficiently disrupt learned patterns. For this reason, 2D spatial dropout, as proposed by Tompson et al. (2015), was applied to achieve regularization by dropping entire feature maps.

### 2.4.4 Weight initialization

The initialization of weights is important for model performance. If weights are initialized with too small values it can lead to the gradients vanishing when the errors are propagated through the model layers, slowing down or halting the learning process. If the weights are set too large, the gradients can grow exponentially leading to exploding gradients. This causes unstable updates of model weights, which in turn results in unstable learning and poor performance. To address these issues, the initialization technique proposed by He et al. (2015) was applied.

### 2.4.5 Activation function

The original U-Net architecture uses the ReLU activation function between convolutional layers. However, an alternative activation that closely resembles the ReLU function called Swish has emerged, with findings indicating that it could outperform ReLU (Ramachandran et al., 2017). Initial tests comparing ReLU and Swish revealed that Swish tended to perform slightly better for the task and data in this study. Consequently, Swish was selected for implementation after each convolutional layer.

For the final layer, the softmax function was applied. This function outputs a vector with a length equal to the number of classes in the dataset, where each number represents the probability of a pixel belonging to a specific class. These probabilities sum to 1 and the pixel is assigned to the class with the largest probability.

### 2.4.6 Loss-function

From initial tests and visual inspections of the results using different loss functions, the focal Tversky loss (eq. 2) after Abraham and Khan (2019) demonstrated the best performance, producing coherent polygons with minimal noise or graininess.

By calculating the Tversky index (eq. 1) independently for each class and averaging over all classes (N), the focal Tversky loss considers the true positives (TP), false positives (FP), and false negatives (FN). The parameters α and β adjust the emphasis of FP and FN, with α adjusting FP and β adjusting FN.

Additionally, the focal parameter (γ) affects the impact of different training examples, which can help mitigate effects of class imbalance (Abraham & Khan, 2019).

$$\text{Tversky index} = \frac{TP}{TP + \alpha \times FP + \beta \times FN} \quad (1)$$

$$\text{Focal Tversky loss} = \frac{\sum_{i=1}^{N} (1 - \text{Tversky Index}_i)^{1/\gamma}}{N} \quad (2)$$

## 2.5 Implementation and training

### 2.5.1 Hardware and software

The model was implemented using the Keras API with TensorFlow backend and trained on a NVIDIA RTX 8000 GPU with 48GB of memory and CUDA capabilities.

### 2.5.2 Model training and hyperparameter optimization

Models were trained for up to 80 epochs, using a batch size of 16 and shuffling the data between epochs. To find hyperparameter values that work well for the network and the given task, Optuna's hyperparameter optimization framework (Akiba et al., 2019) was used together with the logging and visualization capabilities of Weights & Biases (Biewald, 2020). In total, 200 instances of the model were trained and tested using Optuna.

### 2.5.3 Optimization

A search space was defined for key hyperparameters, as outlined in Table 2. The search space included model parameters such as the number of filters, filter size, learning rate, and dropout rate. Additionally, the alpha, beta, and gamma parameters of the focal Tversky loss function were optimized using Optuna.

*Table 2. Hyperparameters and search intervals as input to Optuna.*

| Hyperparameter | Data type | Interval |
| --- | --- | --- |
| **Model parameters** | | |
| Number of filters | Int | [8, 32] |
| Filter size | Int | [3, 7] |
| Learning rate | Float | [0.001, 0.00001] |
| Dropout rate | Float | [0.0, 0.5] |
| **Loss parameters** | | |
| Alpha | Float | [0.3, 0.7] |
| Beta | Float | 1 - alpha |
| Gamma | Float | [1, 3] |

A pruning strategy using Optuna's median pruner was implemented to streamline the hyperparameter optimization. To help achieve this, the first 10 trials were set to run without pruning to establish a baseline. Pruning was then applied to all subsequent trials. The first 30 epochs of each trial were allowed to run without pruning to ensure that models initially performing poorly due to random weight initialization had sufficient time to improve. After the initial 30 epochs, a model was pruned if its performance fell below the median of all previous trials.

## 2.6 Evaluation criteria

### 2.6.1 Confusion matrix and performance metrics

Model evaluation is essential in the training and implementation of DL models, enabling the assessment of model performance during both training and final evaluation. Quantifying agreement between the predicted mask and reference mask was achieved through calculating a population confusion matrix for the final predictions on the test data (Olofsson et al., 2014). The matrix systematically compares reference classes with predicted classes and includes four key outcomes: true positives (TP), negatives (TN), false positives (FP), and false negatives (FN). Due to the large number of pixels, the confusion matrix was normalized by dividing each cell by the total pixel count, so each cell represents a proportion relative to the overall total.

Overall accuracy (OA), as shown in eq. 3, is a common metrics that is often reported, and it is easy to interpret. OA is defined as the correct predictions over all predictions; hence it measures how often the model predictions are correct.

$$\text{Overall accuracy} = \frac{TP + TN}{TP + TN + FP + FN} \quad (3)$$

To provide a more nuanced evaluation, producer's accuracy (PA) given in eq. 4 and user's accuracy (UA) shown in eq. 5 were calculated for each. Producer's accuracy is the ratio between TP and the number of reference pixels in that class (TP + FN), reflecting the models omission errors. User's accuracy measures the ratio between the number of TP for a class and the total number of predicted pixels in that class (TP + FP), indicating commission errors. Producer's and user's accuracy are also commonly known as recall and precision, respectively, in machine learning applications.

$$\text{Producer's accuracy} = \frac{TP}{TP + FN} \quad (4)$$

$$\text{User's accuracy} = \frac{TP}{TP + FP} \quad (5)$$

For unbalanced datasets the accuracy metric tends to give biased estimates. In unbalanced datasets the model will get a large accuracy value simply by predicting the majority class. Another reliable metric for binary classification tasks is the Mathews correlation coefficient (MCC) shown in eq. 6 (Matthews, 1975). MCC considers all correctly (TP, TN) and incorrectly (FP, FN) classified instances. The MCC metric ranges from -1 to 1, where -1 indicates all instances being incorrectly classified, 0 indicates random predictions, and 1 represents perfect classification. MCC is widely regarded as a reliable and robust metric for binary classification tasks (e.g. Chicco & Jurman, 2023).

$$MCC = \frac{TP \times TN - FP \times FN}{\sqrt{(TP + FP)(TP + FN)(TN + FP)(TN + FN)}} \quad (6)$$

To extend MCC to multiclass classification, the metric was calculated individually for each class and then averaged across all classes (N) – a process known as macro averaging. Macro averaging ensures that all classes are weighted equally, regardless of their relative abundance, which is important when all classes are considered equally important in the evaluation. This extended metric is shown in eq. 7 and will be referred to as mMCC.

$$mMCC = \frac{\sum_{i=1}^{N} MCC_i}{N} \qquad (7)$$

### 2.6.2 Visual interpretation

Because of the nature of the delineation process and the fact that the reference data represents only a single realization, the evaluation process cannot fully rely on metrics derived from the confusion matrix. While large values for the accuracy or mMCC metric would suggest good performance, small boundary discrepancies inevitably introduce misclassified pixels. Moreover, as the reference represents only one of many possible versions, these discrepancies do not always indicate actual errors.

The confusion matrix treats all correctly and incorrectly classified pixels uniformly. However, the reality is more nuanced. For instance, a pixel predicted as Class IV but belonging to Class III is arguably less misclassified than if it belonged to Class II, as the former represents a smaller deviation. In other words, the degree of error varies depending on how close the actual class is to the predicted class.

To gain a more detailed understanding of how the model handles the complexities of different stand boundaries, the images were inspected visually. A few representative examples displaying key characteristics of the model predictions were selected and are presented in the results section.

## 3. Results

### 3.1 Tuning and the final model

The training history of the best-performing model configuration is shown in Figure 4. The training loss decreased consistently, showing that the model was successfully minimizing the error on the training set. The validation loss initially decreased rapidly, then showed an overall trend to fluctuate around the training loss, suggesting convergence of the model. This model balanced the alpha and beta parameters of the focal Tversky loss function, and applied a value of 1.3 for the focal parameter (γ), consistent with the value proposed by Abraham and Khan (2019). Additionally, the model applied 3×3 filters, with 29 filters in the initial layer (Figure 3). This configuration achieved maximum validation mMCC and accuracy values of 0.64 and 0.73, respectively, after 73 epochs, and was automatically saved by Weights

& Biases at this stage. When evaluated on the test data, the model achieved mMCC and accuracy values of 0.63 and 0.72.

A more in-depth look at the tuning process revealed that models with smaller filters tended to outperform models utilizing larger filters. Plotting the results of some models using larger filters showed a tendency to produce maps with less noise, but at the cost of more inaccurate boundaries.

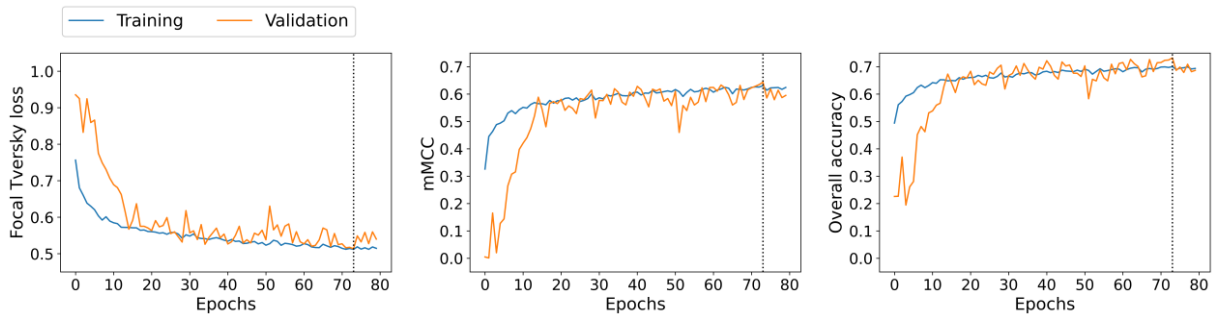

*Figure 4. Training history for the best performing model, showing development of the focal Tversky loss, mMCC, and overall accuracy for both training (blue) and validation (orange) data over 80 epochs. A dotted vertical line marks the best epoch, at which the model was automatically saved by Weights & Biases.*

### 3.2 Classification and confusion matrix

The normalized confusion matrix used to evaluate agreement between the predicted and reference masks is shown in Table 3. The matrix has been normalized by dividing each cell by the overall sum, providing a clear view of the model's performance across different classes.

For Class NF and Class I-II, the model demonstrated strong performance, with UA and PA of 85% and 84% respectively, indicating that the model correctly classified a large proportion of pixels in these classes. The largest source of error for these classes was misclassification as Class III (young thinning phase forest), though misclassification rates to other classes were relatively low.

Class III also showed great accuracy, with a PA and UA values of 73%, effectively balancing commission and omission errors. The main errors for this class were confusion with Class IV (old thinning-phase forest), as indicated by the relatively large misclassification rates between the two classes.

Class IV proved particularly challenging for the model, achieving substantially smaller UA and PA values than the other classes. Similarly to Class III, Class IV showed a balance between UA and PA, indicating that predictions for this class were also balanced between omission and commission errors. The primary source of confusion was classification as either Class III or Class V (forest ready for harvest), with Class V being the largest source of confusion.

Class V had the lowest rate of omission errors, with a PA of 79%. However, the UA value for Class V was notably smaller than the PA, indicating that commission errors were a larger issue. This was primarily caused by confusion with Class IV.

*Table 3. Normalized confusion matrix comparing the predicted mask with the reference masks in the test data. Each cell gives the proportion relative to the total number of pixels being evaluated. The detected and correctly classified pixels (TP) are represented by the bold elements along the diagonal of the shaded area. Producer's accuracy (PA) and user's accuracy (UA) are calculated for each class, giving insights into omission and commission errors.*

|  |  | Reference | | | | | | |
|--|--|--|--|--|--|--|--|--|
|  |  | NF | I-II | III | IV | V | Sum | UA |
| Predicted | NF | **0.12** | 0.00 | 0.01 | 0.01 | 0.01 | 0.14 | 0.85 |
| | I-II | 0.00 | **0.11** | 0.01 | 0.00 | 0.00 | 0.13 | 0.84 |
| | III | 0.02 | 0.02 | **0.21** | 0.03 | 0.01 | 0.29 | 0.73 |
| | IV | 0.00 | 0.00 | 0.05 | **0.10** | 0.02 | 0.18 | 0.55 |
| | V | 0.01 | 0.01 | 0.01 | 0.05 | **0.18** | 0.26 | 0.70 |
| | Sum | 0.15 | 0.15 | 0.29 | 0.18 | 0.23 | 1 | |
| | PA | 0.78 | 0.75 | 0.73 | 0.55 | 0.79 | | |

### 3.3 Visual interpretation and boundary accuracy

Applying the model to the test data showed a fast inference time of just 20 seconds for the entire test dataset, demonstrating an ability to greatly reduce the time consumption for stand delineation.

Visual inspection of the boundaries produced by the model revealed promising results. Example 1, shown in Figure 5, displays the model's prediction and reference for a single image tile. The figure demonstrates that the model can accurately recreate the boundaries defined by the manual interpreter. Close visual inspection of the clear-cut boundaries shows that the model effectively captures the nuances and intricacies of these edges.

Comparing the clear-cut boundaries reveals a discrepancy in border location between the aerial images and CHM. Notably, both the reference and predicted boundaries align with the CHM. Similar results, were the boundaries most closely resemble the CHM, were observed across other images and classes.

For instance, the model shows comparable behavior in delineating forest roads adjacent to mature forests, where tree shadows obscure the edges in images but remain clear in the CHM, potentially indicating a higher reliance on the CHM for accurate delineations.

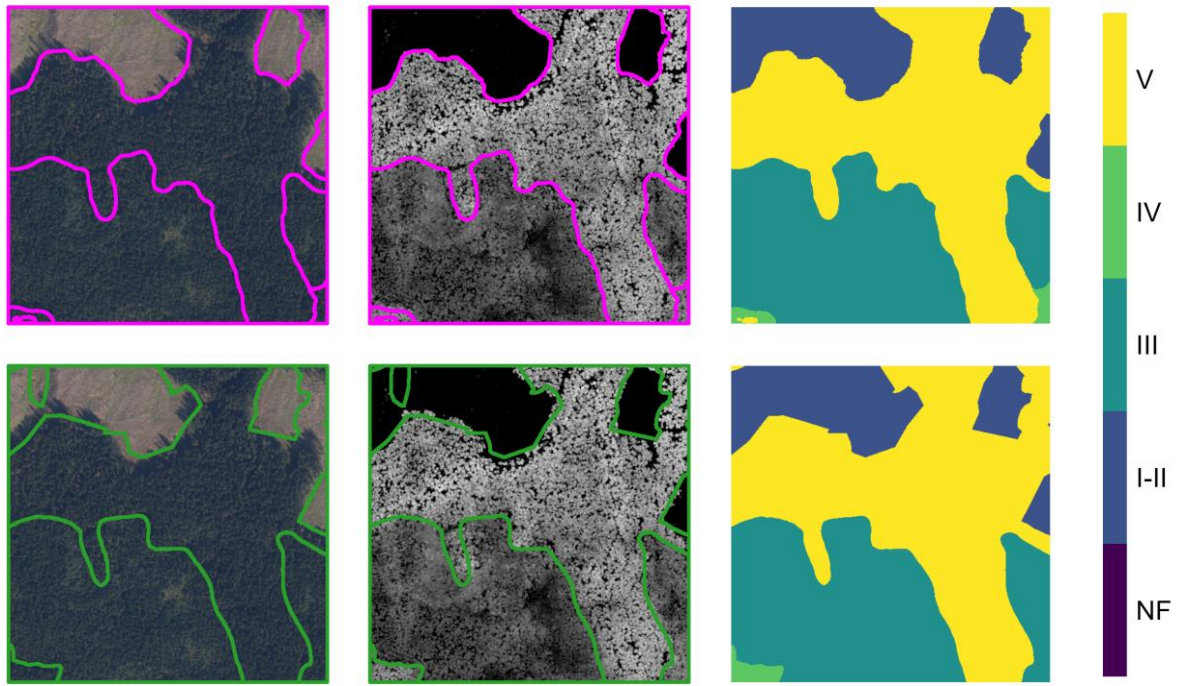

*Figure 5. Model prediction (top) and reference data (bottom) for example 1. The left column shows the boundaries laid over a regular RGB image, the middle column displays the boundaries on top of the CHM, and the right column shows the classification masks. (Classes: NF – non-forest, I-II - V – development stage).*

Example 2, shown in Figure 6, showcases another important aspect of the model predictions. There is good agreement in the overall boundary placement of the stands. However, there are a few misclassified stands (stands 1 and 2), and the model has combined two stands (stands 3 and 4) into a single large stand. According to the forest management plan, stands 1 and 2 are 44 and 43 years old, respectively, and both stands have the same site index value. Based on the definition given by Heje and Nygaard (1999), the lower age limit for classifying these stands as Class IV is 45 years. Therefore, if the age given in the forest management plan is assumed correct, stands 1 and 2 are only one and two years away from qualifying as Class IV, meaning the model's predictions could be considered accurate. Additionally, since the aerial images were acquired one year after the forest management plan was created, stand 2 would already have reached Class IV at the time of image acquisition.

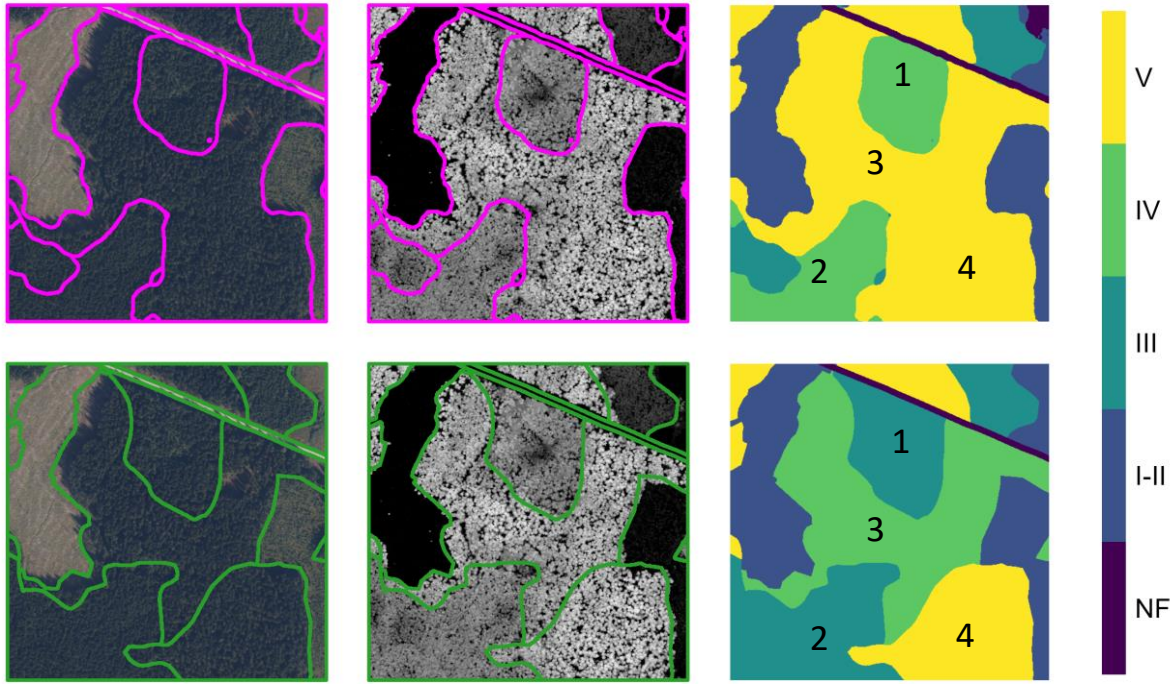

*Figure 6. Model prediction (top) and reference data (bottom) for example 2. The left column shows the boundaries laid over a regular RGB image, the middle column displays the boundaries on top of the CHM, and the right column shows the classification masks. (Classes: NF – non-forest, I-II - V – development stage).*

Comparing the forest management plan inventory and the model predictions, stands 3 and 4 appear similar (Table 4). For stand 3 the lower age limit for Class is 70 years, while the reported age is 67 years, indicating it's close to reaching the predicted class. This similarity suggests that, in terms of forest stand definition, the two stands are homogenous and could be treated as a single operational unit for silvicultural treatment. The goal of delineation is to create operational units, supporting the model's decision to combine these areas.

*Table 4. Stand characteristics for stands 3 and 4.*

| Attribute | Stand 3 | Stand 4 |
| --- | --- | --- |
| Class | IV | V |
| Age | 67 | 82 |
| Site index ($H_{40}$) | G20 | G20 |
| Tree species composition[1] (S, P, B) | (99, 0, 1) | (100, 0, 0) |
| Volume (m³/ha) | 462,0 | 497,2 |
| Height (m) | 23,6 | 24,5 |
| Basal area (m²/ha) | 46 | 48 |
| Mean diameter (cm) | 26,3 | 27,4 |

[1] Percentage distribution of spruce (S), pine (P), and broadleaves (B)

Stand delineation requires considering the entire scene in its context, which poses a challenge for the model. As seen in Example 3 in Figure 7, the model treats a small stand in the top center as a separate unit within the marsh, while the manual interpreter connects it to a nearby stand with similar characteristics. Only considering forest characteristics, the model's boundaries are more accurate, as the manual interpreter includes marshland. However, as the delineation process aims to define operational units, and the smaller stand is too small to function as a separate unit, it is logical to connect the two stands. This pattern is observable across multiple images, where the model separates small regions based on forest characteristics but doesn't account for operational requirements.

Another notable fact in Example 3 is found in the center of the image. The manual interpreter has extended the boundary leftward compared to the model. There is no clearly defined boundary in this region of the image; instead, there is a gradual transition in characteristics, with the stand characteristics of the adjacent stand differing only slightly.

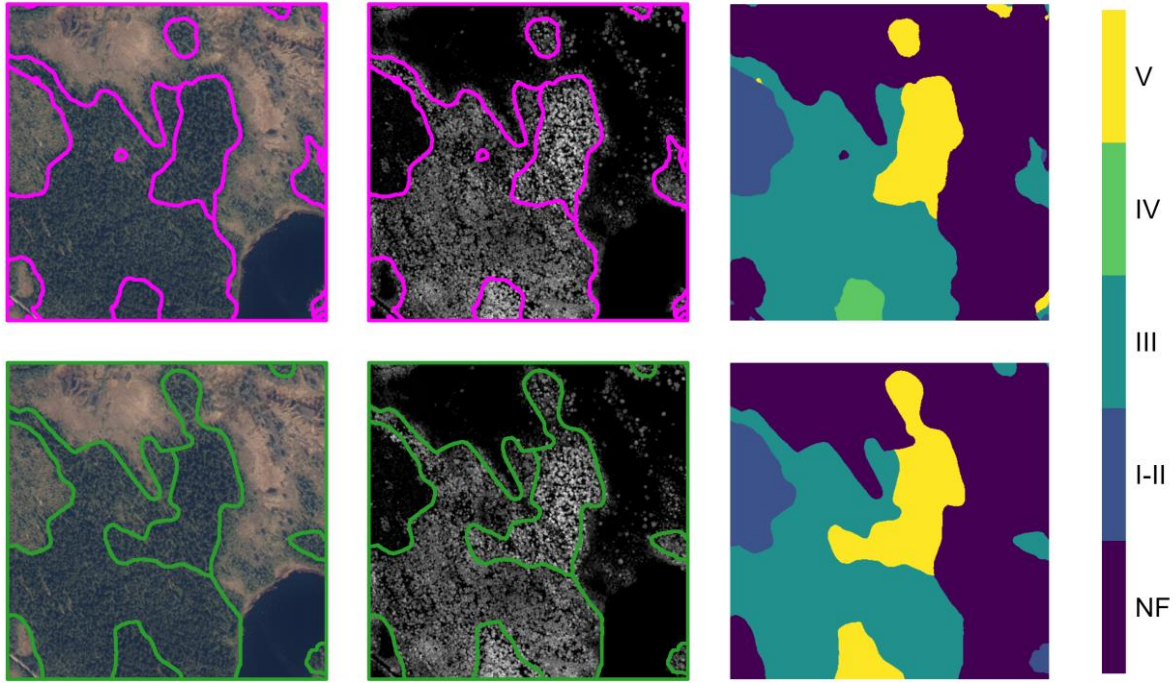

*Figure 7. Model prediction (top) and reference data (bottom), example 3. The left column shows the boundaries laid over a regular RGB image, the middle column displays the boundaries on top of the CHM, and the right shows giving the classification masks. (Classes: NF – non-forest, I-II - V – development stages).*

## 4. Discussion

The model demonstrates solid performance, effectively aligning with the general trends of the reference dataset. The mMCC and overall accuracy indicate strong agreement between the predicted and reference mask for the test data, with values of 0.63 for mMCC and 0.72 for overall accuracy, only slightly smaller than the 0.64 and 0.73 values observed for the validation data during training. These results highlight the model's ability to generalize effectively within the MEV property subject to analysis in this study, which is characterized by intensive silvicultural practices, minimal variability in tree species composition – primarily pure spruce forests – and clear stand structure. Additionally, the proposed DL framework offers rapid inference time, substantially reducing the time and cost associated with manual interpretation. This efficiency is particularly valuable in large-scale operational forestry, for which implementing a tiling strategy could enable the production of continuous stand maps (Huang et al., 2018).

However, while these conditions facilitate strong model performance, they limit generalizability of the findings to less managed forests. For instance, Diedershagen et al. (2004) observed that their automated framework for delineation based on CHM performed worse in structurally heterogenous forest, and was unable to make distinctions between adjacent stands comprising different tree species.

The current study utilized a relatively small dataset of only a few hundred images. While U-Net has demonstrated success with small datasets in certain applications (e.g. Ronneberger et al., 2015) , these results often depend on highly controlled conditions. In contrast, stand delineation is subjective and frequently produced under time pressure, introducing variability that may not be well-captured by the current dataset. Further testing across diverse forest types and stand structures using a larger dataset is necessary to confirm the model's generalizability and ensure robustness.

Expanding the dataset to include a broader range of forest conditions, management practices, and phenological stages would likely enhance the model's performance and its applicability to more diverse contexts. The study found that the model relied heavily on the CHM, potentially due to the structural information in the CHM being particularly important for delineation. However, the one-year time discrepancy between the ALS data and aerial images complicates this interpretation. Another possibility is that this reliance stems from better agreement between the ALS-derived CHM and the reference data. Future work could incorporate a point cloud derived from digital aerial photogrammetry to address temporal inconsistencies, providing more cohesive data inputs and better boundary alignment across input data sources and reference masks.

Another notable source of discrepancy is the model's difficulty with stand boundaries between stands of similar characteristics. This effect would likely also be observed between different interpreters in manual interpretation, but this is not testable in the current study as only the product of a single interpreter is available. Alternatively, the issue could stem from the model lacking sufficient contextual information to make informed decisions. For example, forestry operations are influenced by terrain properties, as they determine accessibility, harvesting feasibility, and operational efficiency (Berg, 1981; Silversides & Sundberg, 1989). In Norway, systems for terrain classification have been made and applied to define operational units (Anon, 1987; Samset, 1975). Manual interpreters often rely on terrain features, such as ridges, valleys, and slopes, to delineate stand boundaries in a way that align with operational requirements. However, this information was absent from the provided model inputs. Incorporating terrain indices, derived from a digital terrain model, could potentially improve the model's ability to delineate boundaries in complex cases and produce outputs that are more in line with operational needs.

The Classes IV and V proved difficult to segment and this led to substantial misclassification issues, suggesting challenges in distinguishing these classes, potentially caused by overlapping feature characteristics. However, after discussing these findings with the expert interpreter it was revealed that this is likely a semantic issue. The instructions used for delineation during the 2021 forest management inventory did not strictly follow the class definitions given by Anon (1987) in all cases, leading to inconsistencies. A similar conclusion was drawn by Tiede et al. (2004) who found that misclassification was primarily related to semantic issues in class definitions, linked to the underlying mapping guidelines.

While the current results are promising, there remains considerable room for improvement. Adaptations of U-Net, incorporating more complex architectures, have demonstrated potential for enhanced performance. For example, combinations with residual networks (Diakogiannis et al., 2020) can help build deeper networks extracting hierarchical features, while inception modules (Cahall et al., 2019) capture features across multiple scales. Implementing attention gates (Oktay et al., 2018) can help the model focus on the most relevant aspects of an image. These strategies have shown to improve segmentation accuracy in other applications, and future research could incorporate these strategies for stand delineation.

As noted by Pukkala (2021), different criteria can guide the delineation process, making direct comparisons between methods challenging. Recent advancements in automated procedures for stand delineation have demonstrated the ability to produce stands that are more internally homogenous compared to manual interpretations (Pukkala et al., 2024; Xiong et al., 2024). However, this is not always the main goal of the delineation process. Historical and cultural silvicultural traditions, along with local conditions, also represent an important frame for the delineation process. Importantly, stands also function as operational units. In this context DL frameworks could be especially useful, as they try to replicate the decisions of the manual interpreter while accommodating the unique criteria, traditions, and local conditions that influence the delineation process.

The definition of "ground truth" poses a key challenge in this study. While the model shows good agreement with the reference dataset, the subjectivity inherent in its creation introduces potential biases. Several factors impact the quality of the reference data, including the limited number of acquisition dates, which restricts the data variability. Spectral signatures change in accordance with phenological changes and can produce quite different impressions depending on the time of acquisition. Additionally, handling spatial data introduces inherent correlations. Despite splitting the dataset into training, validation, and test sets, and excluding the test set during training, nearby images remain spatially correlated. The entire property is also managed under a single silvicultural regime, and the three datasets are all derived from the same ALS and aerial image acquisitions. These factors combined can limit the generalizability of the model. To fully exploit the strengths of DL methods and make models capable of accounting for all aspects of the delineation process, it is essential to extend the dataset to incorporate data from different areas and multiple interpreters. This would improve the robustness of the reference data by capturing a broader range of expert perspectives and delineation conditions, enabling the model to account for all aspects of the delineation process.

## 5. Conclusion

This study presents a novel approach to automated stand delineation and has shown great potential for the implementation of DL algorithms. The model, based on the U-Net architecture, demonstrates strong potential for automating stand delineation in well managed forest environments, aligning well with the reference data and offering substantial efficiency improvements over manual methods. However, its application to more complex and heterogenous forest conditions remains untested, and further research is needed to enhance generalizability. Expanding the dataset, integrating additional data types, improving the alignment of the input data, and exploring more advanced architecture will be essential to improve performance and ensuring that the model can handle diverse forest conditions and different delineation criteria. Ultimately, while the model is a promising step towards automating stand delineation using DL, future work must address the challenges of variability, data complexity, and the subjective nature of "ground-truth" to maximize its applicability and robustness in real-world applications.

## Acknowledgements

We sincerely thank Ngoc Huynh Bao from the Norwegian University of Life Sciences for help getting started with the U-Net model, Mathiesen Eidsvold Værk ANS for providing data, and Viken Skog SA for preparation of data.

## CRediT Authorship contribution statement

**Håkon Næss Sandum:** Conceptualization, Methodology, Visualization, Validation, Writing – original draft; **Oliver Tomic:** Conceptualization, Supervision, Writing – review and editing; **Hans Ole Ørka:** Conceptualization, Supervision, Writing – review and editing; **Erik Næsset**: Funding acquisition, Writing – review and editing; **Terje Gobakken:** Conceptualization, Funding acquisition, Project administration, Supervision, Writing – review and editing.

## Disclosure statement

The authors declare that they have no known competing financial interests or personal relationships that could have appeared to influence the work reported in this paper.

## Data statement

The airborne laser scanning data and the stand map are owned by the forest owner, the aerial images are owned by the Norwegian mapping authority and the authors do not have permission to share the data.

## Funding sources

This work was supported by the Center for Research-based Innovation SmartForest: Bringing Industry 4.0 to the Norwegian forest sector (NFR SFI project no. 309671, smartforest.no).

During the preparation of this work the author used Chat GPT in order to improve readability. After using this service, the author reviewed and edited the content as needed and take full responsibility for the content of the published article.